\documentclass[pdflatex,sn-mathphys-num]{sn-jnl}% Math and Physical Sciences Numbered Reference Style
\usepackage{url}
\usepackage{graphicx}%
\usepackage{multirow}%
\usepackage{amsmath,amssymb,amsfonts}%
\usepackage{amsthm}%
\usepackage{mathrsfs}%
\usepackage[title]{appendix}%
\usepackage{xcolor}%
\usepackage{textcomp}%
\usepackage{manyfoot}%
\usepackage{booktabs}%
\usepackage{algorithm}%
\usepackage{algorithmicx}%
\usepackage{algpseudocode}%
\usepackage{listings}%
\theoremstyle{thmstyleone}%
\theoremstyle{thmstyletwo}%

\theoremstyle{thmstylethree}%

\begin{document}

\title[Article Title]{Forest-Guided Clustering - Shedding Light into the Random Forest Black Box}


\author*[1]{\fnm{Lisa} \sur{Barros de Andrade e Sousa}}\email{lisa.barros@helmholtz-munich.de}

\author[2]{\fnm{Gregor} \sur{Miller}}

\author[2]{\fnm{Ronan} \sur{Le Gleut}}

\author[1]{\fnm{Dominik} \sur{Thalmeier}}

\author[1]{\fnm{Helena} \sur{Pelin}}

\author*[1]{\fnm{Marie} \sur{Piraud}}\email{marie.piraud@helmholtz-munich.de}

\affil[1]{\orgdiv{Helmholtz AI}, \orgname{Helmholtz Munich}, \orgaddress{\street{Ingolstädter Landstraße 1}, \city{Neuherberg}, \postcode{85764}, \state{Bavaria}, \country{Germany}}}

\affil[2]{\orgdiv{Core Facility Statistical Consulting}, \orgname{Helmholtz Munich}, \orgaddress{\street{Ingolstädter Landstraße 1}, \city{Neuherberg}, \postcode{85764}, \state{Bavaria}, \country{Germany}}}


\abstract{As machine learning models are increasingly deployed in sensitive application areas, the demand for interpretable and trustworthy decision-making has increased. Random Forests (RF), despite their widespread use and strong performance on tabular data, remain difficult to interpret due to their ensemble nature. We present Forest-Guided Clustering (FGC), a model-specific explainability method that reveals both local and global structure in RFs by grouping instances according to shared decision paths. FGC produces human-interpretable clusters aligned with the model’s internal logic and computes cluster-specific and global feature importance scores to derive decision rules underlying RF predictions. FGC accurately recovered latent subclass structure on a benchmark dataset and outperformed classical clustering and post-hoc explanation methods. Applied to an AML transcriptomic dataset, FGC uncovered biologically coherent subpopulations, disentangled disease-relevant signals from confounders, and recovered known and novel gene expression patterns. FGC bridges the gap between performance and interpretability by providing structure-aware insights that go beyond feature-level attribution.}




\maketitle

\section{Introduction}\label{sec1}

In recent years, machine learning (ML) models have demonstrated remarkable performances across many application areas, ranging from medical diagnostics \citep{medical_diagnostics_2023} and financial fraud detection \citep{fraud_detection_2024} to smart grid load forecasting \citep{smart_grid_2024}. Complex ML models are frequently referred to as "black boxes" due to the difficulty in understanding the reasoning behind their predictions. As ML models are increasingly deployed in critical areas such as healthcare, finance, human resources or energy management, understanding \textit{how} and \textit{why} these systems make decisions has become a major concern. As stated by Doshi-Velez et al. \citep{doshi_2017}: “The problem is that a single metric, such as classification accuracy, is an incomplete description of most real-world tasks”. In other words, accuracy alone cannot capture essential modelling aspects such as fairness, robustness, or accountability, especially in domains where understanding the \textit{why} behind a prediction is as important as the prediction itself. The field of eXplainable Artificial Intelligence (XAI) addresses this challenge by aiming to improve the transparency of complex models to ensure that their decisions can be interpreted, trusted, and audited by humans. This need also stems from the European Union’s General Data Protection Regulation (GDPR), which includes a “right to explanation,” forcing organizations to provide meaningful information about automated decision-making processes that significantly affect individuals \citep{gdpr1, gdpr2}. 

The importance of integrating XAI analysis into modeling workflows is further highlighted by various real-world failures of deployed AI systems, where the lack of explainability has resulted in biased, unfair, and sometimes harmful predictions. In healthcare, a large-scale study uncovered that a widely used risk prediction algorithm, intended to identify patients for high-risk care management, systematically underestimated the needs of black patients due to wrong modeling assumptions. The algorithm predicted healthcare costs instead of illness as a proxy for health status, ignoring structural inequalities in healthcare access. As a result, black patients with similar levels of illness as white patients received fewer resources \citep{healthcare_bias_2019}. Another example involves a machine learning model developed to predict pneumonia risk. While the model outperformed medical experts in accuracy, it incorrectly identified asthma patients as having a lower risk of death from pneumonia. This counterintuitive result was driven by a data bias, as asthma patients typically receive more intensive care, leading to improved outcomes \cite{pneumonia_model, pneumonia_review}. Biases in training data and the lack of transparency in machine learning models have also led to negative consequences in sectors such as public administration, criminal justice, and human resources. In the U.S., the COMPAS algorithm, used to guide sentencing and bail decisions, was shown to assign disproportionately higher risk scores to black defendants compared to white defendants with similar profiles, raising serious concerns about fairness and accountability in algorithmic decision-making \citep{compas1, compas2}. In the UK, several AI systems used in public services like visa applications and welfare fraud detection, have been shown to disadvantage individuals based on race or nationality, raising questions about transparency and accountability in government use of AI \citep{uk_ai_benefits, uk_ai_visa}. Similarly, Amazon shut down an AI-powered recruitment tool after it was discovered that the model systematically downgraded resumes containing the word "women’s,", reinforcing historical gender biases \citep{amazon_bias}. These cases illustrate the need for explainable AI methods. Without tools that allow humans to understand and evaluate a model’s behavior, we risk loosing public trust, reinforce social inequalities, and fail to meet legal and regulatory requirements.

The field of XAI differentiates between interpretability, which refers to the intrinsic transparency of a model, and explainability, which refers to post-hoc techniques that provide insights into the workings of black-box models like Random Forests (RF). RFs, an ensemble learning method built from single decision trees, have been widely adopted across different domains. \citep{rf_application, rf_application2, rf_application3, rf_application4, rf_application5, rf_application6, rf_application7, rf_application8, rf_application9, rf_application10}. They are particularly suited for high-dimensional tabular datasets, common in many real-world applications, where the number of features may exceed the number of observations. In such scenarios, RFs frequently outperform deep learning models, which typically require large amounts of data and extensive hyperparameter tuning to achieve comparable results \citep{rf_vs_dl}. Their robustness against overfitting comes from the use of bootstrapping and random feature selection, both of which increase diversity among the individual decision trees and enhance generalizability. However, although they consist of individually interpretable trees, the aggregated ensemble structure of RFs reduces the overall model transparency, turning them into a black-box model. Standard model-agnostic XAI methods like Permutation Feature Importance \citep{breiman2001random}, SHAP \citep{lundberg2017unified}, and LIME \citep{ribeiro2016should} are widely used for RF explainability by pinpointing the individual contributions of features to the model output, but they also come with notable limitations. Permutation Feature Importance, also known as Mean Decrease in Accuracy, can be unstable in the presence of correlated features, often resulting in unreliable rankings and distributed importance scores across those correlated features. LIME, on the other hand, is less sensitive to feature correlations but depends on local approximations, which can produce inconsistent explanations for similar inputs and offers limited insights into the model's global behavior. SHAP is among the most advanced methods for explaining RFs, offering both local and global interpretability. However, it struggles to handle complex feature correlations due to its assumption of feature independence and can become computationally expensive for large datasets or deep models. Although widely adopted, such post-hoc XAI methods often fail to capture the influence of correlated features, complex feature interactions, and nonlinear relationships by considering or interpreting features in isolation, missing synergistic interactions that significantly influence predictions. 

To overcome the limitations of model-agnostic explainability techniques several model-specific approaches have been developed for interpreting RFs. Among the earliest was Breiman’s Mean Decrease in Impurity, which quantifies a feature's importance by averaging the decrease in node impurity (e.g., Gini index) caused by splits on that feature across all trees \citep{breiman2001random}. Although widely used, it is biased towards features with many categories or continuous values. More recent methods exploit the model’s internal proximity matrix, first proposed by Breiman \citep{breiman2008manual}, to quantify proximity between instances based on the frequency with which they fall into the same terminal nodes across all decision trees. In unsupervised settings, proximity-based approaches have been used for clustering and visual exploration. For example, Shi and Horvath \citep{shi2006unsupervised} used proximities derived from unsupervised RFs trained on pseudo-labeled data to define a distance matrix, which was subsequently used for clustering, while Shi et al. \citep{shi2005tumor} used a similar approach to identify subtypes in renal carcinoma microarray profiles. In supervised settings, RF proximity measures have been used for dimensionality reduction and data exploration. Pang et al. \citep{pang2006pathway} constructed class-vote vectors for biological pathways and applied multidimensional scaling (MDS) to visualize inter-pathway relationships. Related, Pouyan et al. \citep{pouyan2016distance} derived a proximity matrix from correctly classified instances to learn a class-discriminative distance metric, which was subsequently used for improved classification and visualization of cytometry data. Proximity-based feature attribution has been explored by Whitmore et al. \citep{whitmore2018explicating}, who proposed a method that computes differences in proximity matrices before and after feature permutation to identify influential features and local contributions, combining aspects of permutation importance with LIME.

Our method, called Forest Guided Clustering (FGC), builds on this previous work by extending the proximity-based analysis with the integration of a supervised learning objective into the clustering process. Rather than approximating the model with simpler surrogates or ranking the features through permutation tests, FGC leverages the internal structure of RFs to group data points based on their proximity in tree traversal patterns. This approach offers a human-interpretable segmentation of the dataset, shedding light on how the RF model views the data. By analyzing these forest-guided clusters, users can identify sub-populations in the data, detect biases, and generate domain-relevant hypotheses. This paper introduces the theoretical foundations of FGC, shows practical applications, and compares to state-of-the-art XAI methods, aiming to bridge the gap between model accuracy and model explainability. \\

\section{Results}\label{sec2}

\begin{figure}[ht]
    \centering
    \includegraphics[width=\textwidth]{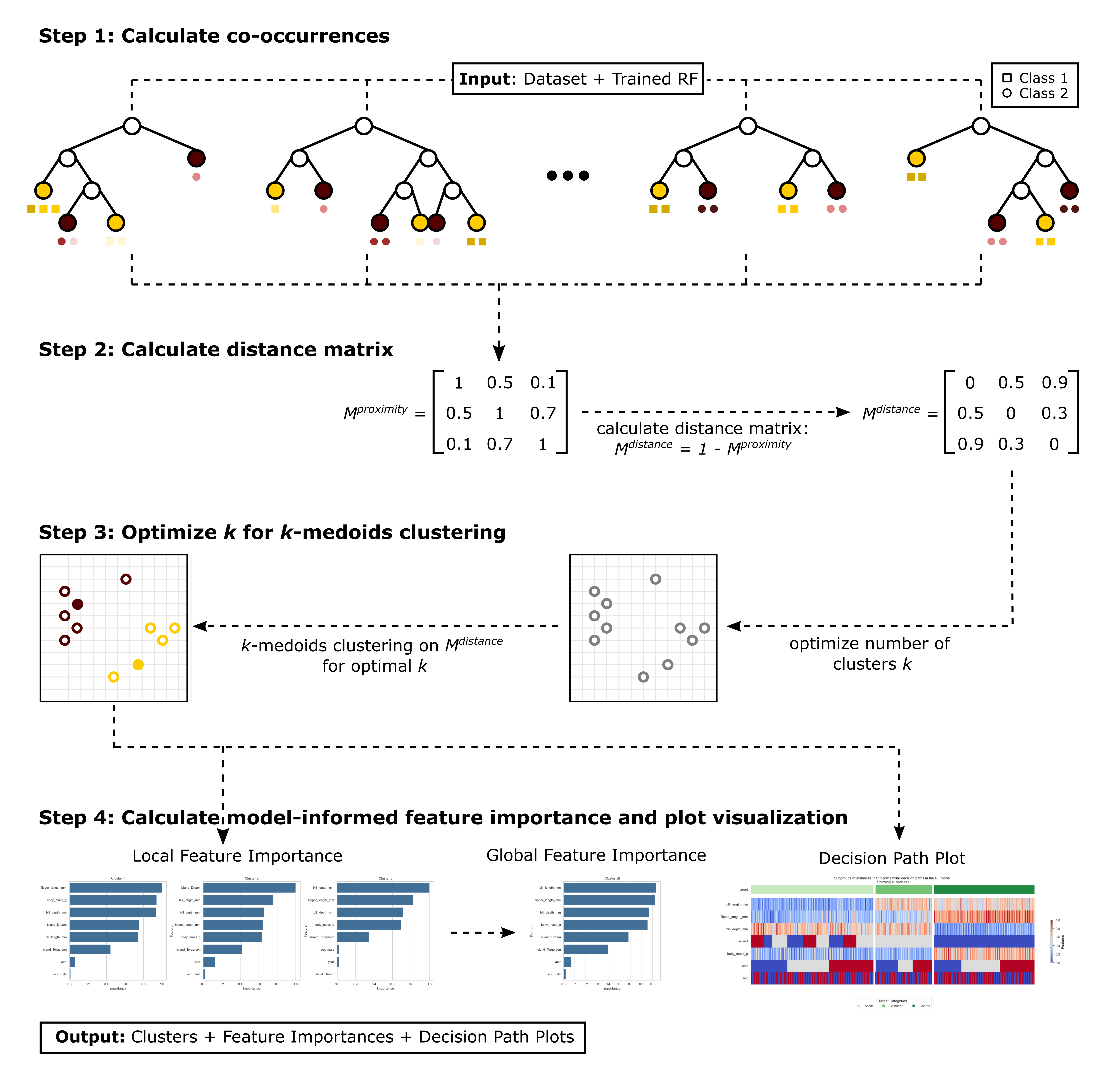}
    \caption{\textbf{Workflow of the Forest Guided Clustering (FGC) algorithm.}
    FGC begins by computing a proximity matrix from a trained RF model, where each entry represents how frequently two samples appear in the same terminal node across all trees. This proximity matrix is then converted into a distance matrix, which is used as input for $k$-medoids clustering. The optimal number of clusters $k$ is determined by jointly minimizing cluster bias and cluster variance. Once clustering is completed, FGC computes local feature importance scores for each cluster, as well as a global importance score by averaging local importance scores across clusters. A decision path plot provides a visual summary of feature enrichment patterns across clusters, facilitating interpretation of the model's learned decision rules.}
    \label{fig:fgc_workflow}
\end{figure}

FGC is a model-specific explainability approach that reveals both, local and global structures in RF models by identifying groups of instances that follow similar decision paths throughout the ensemble. By leveraging the internal structure of the RF, specifically, the tree traversal patterns, FGC derives a segmentation of the input space that reflects the decision rules learned by the RF model, offering interpretable insights into how the model structures the input space. An overview of the FGC algorithm is shown in Figure~\ref{fig:fgc_workflow}, illustrating the main steps from proximity-based distance calculation and clustering, to the computation of feature importance and visualization of the decision path.

The first step of the FGC algorithm is the computation of a proximity matrix that quantifies how often pairs of instances co-occur in the same terminal nodes across the ensemble of trees. This proximity matrix reflects the decision structure learned by the model and is transformed into a distance matrix that captures model-informed dissimilarities (see Section ~\ref{sec4_1}). Pairs of instances that frequently co-occur in the same terminal nodes are interpreted as similar with respect to the learned decision rules. To identify coherent subgroups of instances that are aligned with the RF's decision logic, a $k$-medoids clustering is applied directly to the model-derived distance matrix. Unlike traditional geometric distances, this RF-informed distance captures nonlinear interactions and hierarchical relationships, enabling the clustering to operate in a space defined by the model’s behavior rather than the raw feature input. A critical step of FGC is the selection of the number of clusters, which directly influences the resolution and interpretability of the resulting segmentation. To identify meaningful subpopulations with distinct predictive profiles, while avoiding unnecessary complexity, FGC uses a dual-objective criterion that jointly optimizes cluster bias and cluster variance (see Section ~\ref{sec4_2}). The first objective, cluster bias, quantifies the degree to which cluster assignments align with the target variable, using a class-balanced impurity score for classification tasks or within-cluster variance for regression tasks. The second objective, cluster variance, measures the stability of the clustering under bootstrapped perturbations, using the averaged Jaccard similarity. Only clustering solutions that meet a minimum stability threshold are considered valid, and among these, the one with the lowest bias is selected. This ensures that the final segmentation is both reproducible and aligned with the model's learned structure. Once the segmentation is established, FGC computes local and global feature importance scores (see Section ~\ref{sec4_3}). Local feature importance is calculated by comparing the distribution of feature values within each cluster to a background distribution aggregated across all clusters, using divergence measures such as the Wasserstein or Jensen–Shannon distance. This allows the identification of features that are distinctive for a given cluster and helps to distinguish between features that meaningfully contribute to the formation of a specific cluster versus those that are uniformly distributed and thus less relevant for segmentation. Global feature importance is obtained by averaging local importance scores across clusters, providing a ranked list of features that most strongly differentiate between decision regions learned by the RF. To facilitate interpretability, FGC provides visualization tools that reveal the decision patterns driving cluster formation. The decision path plot combines two complementary views: a feature heatmap and cluster-specific distribution plots. The heatmap provides a high-level overview of feature patterns across clusters, showing the average standardized feature values and ranking features by global importance. Feature enrichment or depletion within clusters is shown through color intensity, allowing users to  identify features that distinguish between subgroups and their alignment with the target distributions. This view also summarizes the target attribution across clusters, helping to identify mislabeled instances or confounding factors. The feature distribution plots display the raw, unstandardized feature values per cluster, allowing for inspection of within-cluster and inter-cluster variation in the original feature space. Together, these visualizations help to understand which general decision rules have been learned across the different trees in the ensemble and which features drive the separation of subpopulations.

The Forest Guided Clustering algorithm is available as an open-source Python package at \url{https://github.com/HelmholtzAI-Consultants-Munich/fg-clustering}, providing a user-friendly implementation for RF model interpretability. In addition, the repository includes several tutorials that demonstrate how to apply FGC in practice, including guidance on scaling the method to large datasets.

\subsection*{Benchmarking FGC Against Clustering and Explanation Methods}

To evaluate how FGC captures both local and global model behavior, we applied it to a simulated dataset designed to reflect a complex but interpretable structure. The dataset consists of two class labels, with informative features encoding a four-subclass structure and two additional noise features unrelated to the target (see Section~\ref{sec4_4}). Although no single informative feature is sufficient to fully separate the classes, their joint distribution enables clear subclass separation. This setup provides a controlled setting to compare FGC against both clustering and explainability baselines (Figure~\ref{fig:simulated_data}).

\begin{table}[ht]
\centering
\begin{tabular}{lccccc}
\toprule
\textbf{Method} & \multicolumn{2}{c}{\textbf{Clustering}} & \multicolumn{3}{c}{\textbf{Feature Importance (FI)}} \\
                & Supervised & Unsupervised & Global & Local & Individual \\
\midrule
\textit{Clustering methods} \\
FGC                           & \checkmark &           & \checkmark & \checkmark &           \\
Unsupervised RF clustering    &           & \checkmark &        &           &           \\
k-medoids clustering    &           & \checkmark &        &           &           \\
\midrule
\textit{Feature importance methods} \\
Mean Decrease in Impurity FI  &           &           & \checkmark &           &           \\
Permutation FI                &           &           & \checkmark &           &           \\
TreeSHAP                      &           &           & \checkmark &           & \checkmark \\
LIME                          &           &           &           &           & \checkmark \\
\bottomrule
\end{tabular}
\caption{\textbf{Comparison of clustering and feature importance (FI) methods.} 
Global FI refers to model-wide importance, local FI to subgroup importance, and individual FI to specific predictions. Note that some implementations refer to individual FI as local FI.}
\label{tab:other_methods}
\end{table}

\begin{figure}[ht]
    \centering
    \includegraphics[width=\textwidth]{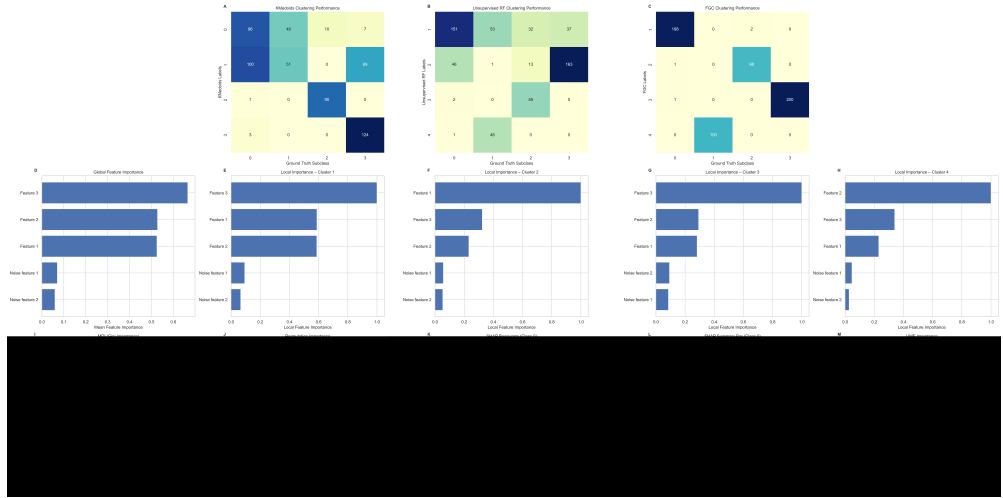}
    \caption{\textbf{Clustering heatmaps and feature importance of different methods.}
    (A-C) Overlap of clusters identified by k-medoids, unsupervised RF, and FGC. (D) Global feature importance of the FGC. (E-H) Local feature importances of the individual clusters identified by the FGC. (I) MDI feature importance of the RF classification. (J) Permutation feature importance of the RF classification. (K) Beeswarm plot with one dot for each instance and feature indicating the SHAP value and the colour indicating the original value of the feature. (L) Mean absolute value of the SHAP values per feature. (M) Mean LIME value per feature.}
    \label{fig:feature_importance_comparison}
\end{figure}

We benchmarked FGC against several established clustering and feature importance methods with publicly available implementations (Table~\ref{tab:other_methods}). For clustering, we benchmarked FGC  against unsupervised $k$-medoids and unsupervised Random Forest (RF) proximity clustering (see Section~\ref{sec4_4} for details). In contrast to FGC,  Both methods disregard class labels, while FGC incorporates class supervision and leverages decision-path similarity to discover latent structure. This label-guided clustering enabled FGC to  identified all four subclasses, achieving an adjusted Rand index (ARI) of 0.98 between predicted clusters and ground-truth subclass labels (Figure~\ref{fig:feature_importance_comparison}A). In contrast, both unsupervised $k$-medoids and RF proximity clustering yielded lower agreement with the ground truth, with ARIs of 0.35 and 0.33, respectively, indicating limited alignment with the underlying data structure (Figure~\ref{fig:feature_importance_comparison}B-C).

We next compared FGC to feature importance methods including model-specific approaches (MDI) and model-agnostic alternatives (MDA, SHAP, LIME). Unlike these methods, FGC produces both global and local feature importance scores, revealing how informative features contribute within and across clusters (Figure~\ref{fig:feature_importance_comparison}D–H). All three informative feature were correctly identified as relevant and Feature 3 was correctly identified as the most globally influential feature, as it governs the decision boundaries in the two largest clusters. While MDI and MDA (Figure~\ref{fig:feature_importance_comparison}I–J) also highlight Feature 3 globally, they do not capture local variation or the underlying decision structure. Similarly, SHAP and LIME offer fine-grained per-instance attributions that can be aggregated (Figure~\ref{fig:feature_importance_comparison}K–M), but do not group samples or expose higher-order organization.

\begin{figure}[ht]
    \centering
    \includegraphics[width=\textwidth]{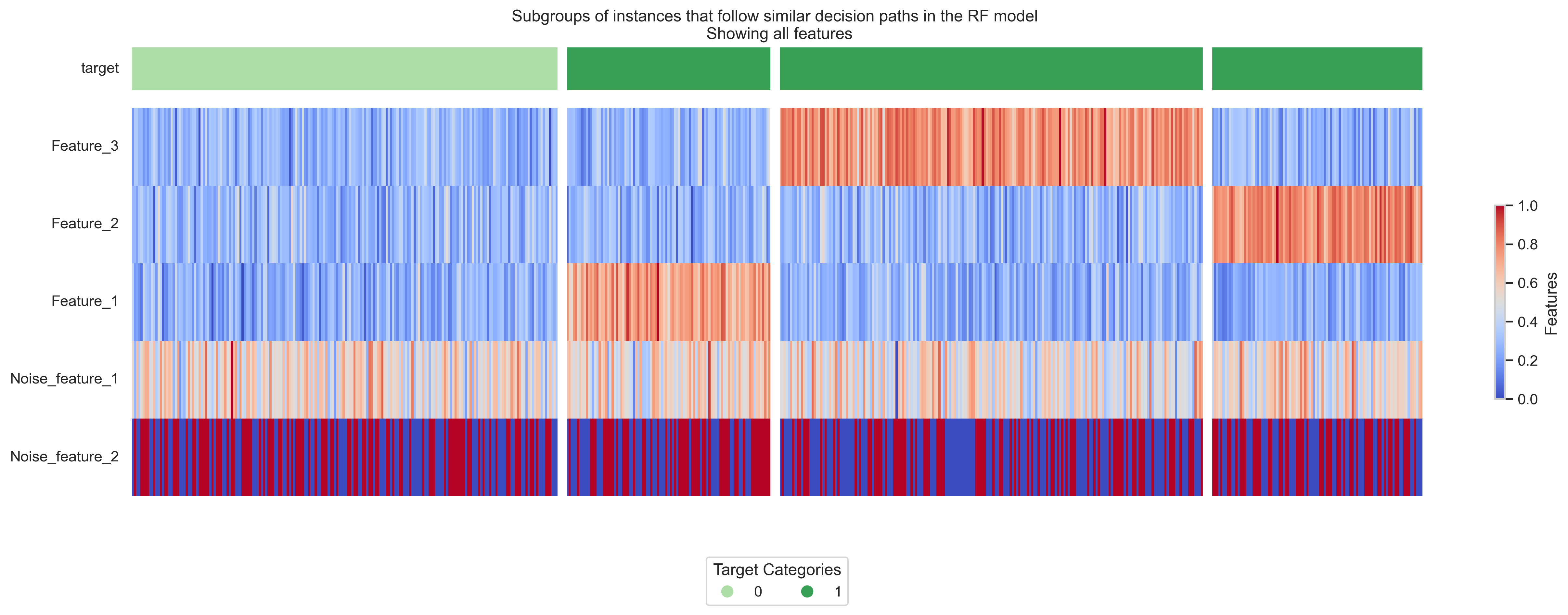}
    \caption{\textbf{Decision paths of FGC for simulated data.}
    The heatmap indicates that FGC was able to identify the three subclasses of class 1, while leaving class 0 as one separate cluster. The noise features did not affect the clustering.}
    \label{fig:simulated_data_heatmap}
\end{figure}

In contrast, FGC clusters samples based on shared decision paths, revealing model-inferred subpopulations. The corresponding heatmap (Figure~\ref{fig:simulated_data_heatmap}) visualizes consistent feature activation patterns across clusters and highlights how specific features guide classification decisions. This integrated view offers a richer model interpretation than traditional attribution methods alone.

\subsection*{Acute Myeloid Leukemia (AML) Case Study}

Acute Myeloid Leukemia (AML) is a highly aggressive hematologic malignancy initiated by genetic alterations in myeloid progenitor cells \citep{hartmann2019clonal}. These mutations lead to the uncontrolled proliferation of undifferentiated myeloid precursors in the bone marrow and peripheral blood, ultimately resulting in severe clinical outcomes like anemia or haemorrhage \citep{de2016acute}. AML shows substantial heterogeneity across patients, driven by a combination of genetic, epigenetic, and clinical factors \citep{li2016genetic, bonnet2005cancer}. In this case study, we leverage a curated RNA-seq dataset spanning over 12,000 genes and a diverse cohort of individuals, including patients with AML, other leukemia types, non-leukemia diseases, and healthy controls \citep{warnat2020scalable}. Dimensionality reduction (Figure~\ref{fig:AML_case_study_dataset}) revealed that sample structure was strongly influenced by study origin, sample tissue type, and disease label, highlighting the presence of potential technical and biological confounders. To mitigate these effects, we trained a RF model for binary AML classification using a cross-study evaluation strategy, which ensured strict separation between training and test samples by study (see Section~\ref{sec4_5} for details). The final model achieved $96.6\%$ balanced accuracy on an independent test set, demonstrating strong predictive performance and robustness to study-specific variation. Our goal in this case study is threefold: (i) apply FGC to reveal how the trained model internally segments the gene expression space based on decision path proximity, (ii) interpret the model’s internal structure by identifying genes that distinguish decision regions, and (iii) assess whether potential confounding factors such as tissue source or study origin influence model behavior.

\begin{figure}[ht]
    \centering
    \includegraphics[width=\textwidth]{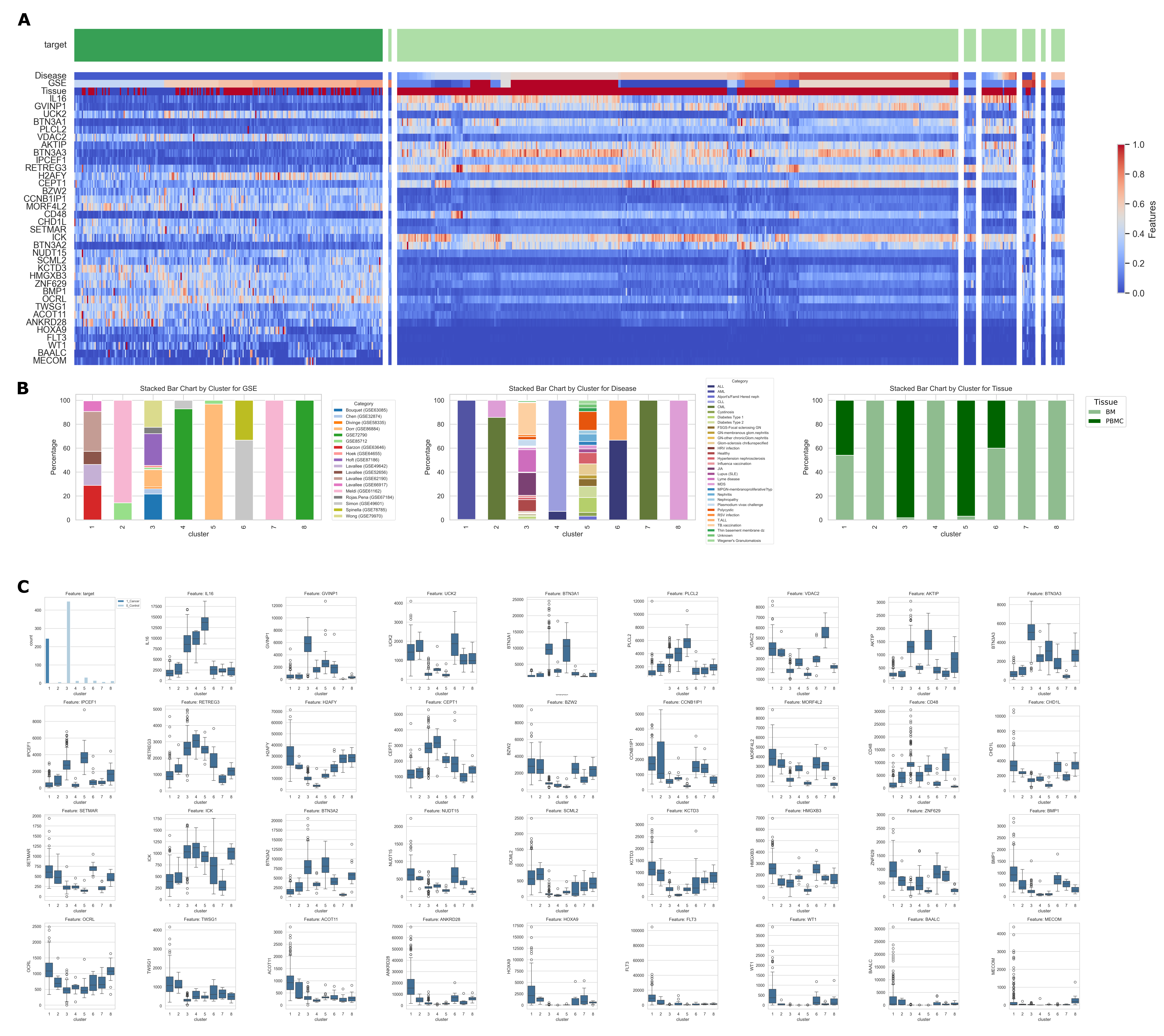}
    \caption{\textbf{Decision path visualization for the AML case study.}
    (A) Cluster-level heatmap showing standardized feature values of the 30 most important genes for the AML-enriched cluster, extended with five genes known to be overexpressed in AML (BAALC, FLT3, HOXA9, MECOM, WT1). Samples are grouped by their FGC cluster assignments. Metadata annotations were added as first rows in the heatmap. 
    (B) Cluster composition by study origin (GSE), disease label, and tissue type. 
    (C) Distribution plots of log$_2$-transformed, normalized gene expression values across clusters for all selected genes.}
    \label{fig:AML_decision_path}
\end{figure}

To interpret the model’s internal decision structure, we applied FGC, which identified eight stable clusters of samples with similar decision paths through the RF (Table~\ref{tab:FGC_cluster_stability}). Local feature importance scores revealed distinct gene expression signatures associated with each cluster (Figure~\ref{fig:AML_case_study_feature_importance}). To further investigate the model’s decision logic and assess potential confounding factors, we visualized decision paths using a targeted gene panel composed of the top 30 genes from the AML-enriched cluster, complemented by five AML-associated genes (BAALC, FLT3, HOXA9, MECOM, WT1) \citep{handschuh2019not}. Cluster composition (Figure~\ref{fig:AML_decision_path}A) revealed clear separation of biological subtypes: Cluster 1 contained only AML samples, Cluster 2 was enriched for CML and some MDS samples, Cluster 4 for CLL samples Cluster 7 for CML samples and Cluster 8 for MDS samples. In contrast, Clusters 3 and 5 comprised non-leukemic conditions (Table~\ref{tab:FGC_disease_distribution}). These results show that, despite the binary classification task, the model internally stratified samples into biologically coherent subgroups. Metadata inspection indicates that the tissue type from which the experimental samples were obtained was not a major confounding factor, as both AML- and ALL-enriched clusters included a mix of bone marrow (BM) and peripheral blood mononuclear cells (PBMCs) samples (Figure~\ref{fig:AML_decision_path}B). However, potential batch effects linked to study origin were evident. For instance, Cluster 5 consisted almost exclusively of control samples from a single study, whereas Cluster 3 aggregated control samples from multiple studies (Table~\ref{tab:FGC_GSE_distribution}). This separation likely reflects study-specific technical variation. FGC helps to identify such effects by uncovering model-defined subpopulations, supporting a systematic evaluation of whether learned decision boundaries reflect underlying biological signals or confounding factors.

To evaluate the biological relevance of model-derived features, we first verified that the FGC recovered known AML markers. Although these genes showed moderate upregulation in the AML-enriched cluster (Cluster 1), confirming that the model recovers established AML signatures (Figure~\ref{fig:AML_decision_path}C, they were not among the top-ranked features. This motivated a deeper investigation of the features prioritized by the model. In multi-disease datasets, gene expression signals can be confounded by differences in cellular origin rather than reflecting malignancy-driven dysregulation. Since hematologic malignancies arise from distinct stages of blood cell development, ranging from immature myeloid progenitors (AML) and hematopoietic stem cells (CML, MDS) to lymphoid precursors (ALL) or mature B cells (CLL), classification features may reflect developmental lineage rather than disease-specific dysregulation. Accounting for this context is essential to distinguish true biological signals from cell-of-origin effects.

To disentangle these factors, we functionally categorized the top 30 model-derived features using UniProt gene ontology annotations grouping them into immune response and cytokine activity, chromatin and epigenetic regulation, transcriptional regulation, catalytic activity, and other functions \citep{bateman2024uniprot}. We then assessed how the expression of these genes varied across FGC-defined clusters to determine whether patterns reflected disease-specific changes or lineage-associated variation. Genes related to immune response and cytokine signaling (e.g., BTN3A1/2/3, CD48, IL16, GVINP1) showed high expression in PBMC-rich control clusters and reduced expression in AML, CML, and MDS. This pattern is consistent with their established roles as immune cell markers, suggesting that immune cell composition rather than leukemia-specific dysregulation is captured. In contrast, chromatin-associated genes (MORF4L2, H2AFY, CHD1L, SETMAR) exhibited elevated expression in AML and other myeloid malignancies, supporting a possible role in epigenetic reprogramming or transcriptional deregulation. Similarly, transcriptional regulators such as ZNF629, BZW2, HMGXB3, and SCML2 showed cluster-specific expression patterns aligning with leukemic types. Features involved in catalytic activity revealed a mixed pattern: genes such as IPCEF1, ICK, and PLCL2 were more highly expressed in controls or lymphoid malignancies, while others, including UCK2, NUDT15, ACOT11, CCNB1IP1, and BMP1, were enriched in AML and CML clusters, in line with their known or proposed oncogenic roles. Lastly, a set of less well-characterized genes (AKTIP, RETREG3, VDAC2, TWSG1, KCTD3, ANKRD28) included both lineage-associated and AML-specific expression patterns. Notably, genes like ANKRD28 and TWSG1 showed AML-driven upregulation, suggesting a potential involvement in disease-associated pathways. Full gene-level expression profiles and corresponding literature references are provided in Appendix~\ref{secA2} and Supplementary Table~\ref{tab:gene_level_summary}.

Together, these analyses show that the top model-derived features not only recovered known AML-associated signatures but also revealed distinct gene groups whose expression is potentially shaped by either malignant transformation or cell-of-origin effects. Immune-related features primarily reflected variation in immune cell composition across hematopoietic lineages, whereas chromatin regulators, transcriptional factors, and catalytic enzymes showed stronger associations with AML, suggesting functional roles in leukemogenesis. Notably, chromatin remodelers such as H2AFY and SETMAR were more highly expressed in AML than CML, highlighting subtype-specific regulatory patterns. Shared upregulation of several chromatin-associated genes in AML and ALL may reflect common features of acute leukemias, including rapid proliferation and aggressive progression. In contrast, CLL displayed reduced expression of proliferation- and chromatin-related genes, with elevated levels of immune-related features, consistent with its more mature lymphoid phenotype and slower disease kinetics. MDS, as a pre-leukemic condition, showed gene expression patterns more similar to control samples than to AML or CML. These findings highlight the ability of FGC to uncover biologically meaningful structure within high-dimensional transcriptomic data, even though the leukemia type differentiation was not explicitly defined by the binary classification objective.

\section{Discussion}\label{sec3}

As machine learning models become increasingly integrated into scientific workflows and critical infrastructure, the need for transparent and trustworthy decision-making is more urgent than ever. In sensitive application domains such as healthcare, finance or energy management, accuracy alone is not sufficient. Understanding why a model makes a prediction is essential for ensuring fairness, accountability, and responsible deployment. Failures of black box AI systems across healthcare, criminal justice, and public administration have demonstrated that a lack of explainability can reinforce biases, undermine trust, and cause real-world harm. As a consequence, explainability methods provide a valuable support in validating model behavior and revealing whether predictions are driven by meaningful signal or confounding factors. Forest-Guided Clustering (FGC) addresses this challenge by complementing traditional feature-centric explanations with a structure-aware global perspective. Instead of attributing importance to individual features in isolation, FGC clusters samples based on shared decision paths and identifies locally relevant features within each group, producing an interpretable segmentation of the input space. This approach offers insight into both individual predictions and the broader decision logic of the model.

The following key aspects distinguish FGC from established explainability methods such as MDI, MDA, and SHAP:

\begin{itemize}
    \item \textbf{Beyond feature independence assumptions:} FGC does not rely on the assumption of feature independence, a common limitation of model-agnostic approaches that can compromise explanatory validity in high-dimensional, correlated datasets. Instead, FGC leverages the structure of the RF model to capture feature interactions and nonlinear decision rules as they are learned.
    
    \item \textbf{Recovery of higher-order structure in model behavior:} Compared to SHAP, which offers detailed per-instance attributions but lacks the ability to group samples into higher-order structure, FGC offers a structured interpretation of the model behavior. It reveals how the model partitions the decision space and what features differentiate classes across subpopulations. This enables the detection of higher-order structure that remains hidden in traditional feature-centric explanation methods.
    
    \item \textbf{Task-guided clustering for label-relevant structure:} FGC also outperforms unsupervised clustering approaches such as k-medoids or unsupervised Random Forests by incorporating class label information. This enables FGC to uncover label-relevant subgroups and detect structure aligned with the modeling objective.
    
    \item \textbf{Scalability to large datasets:} To ensure practical applicability, FGC is provided as a scalable version using the CLARA algorithm and efficient matrix computations, which enables clustering and feature analysis on datasets with thousands of samples.
\end{itemize}

Taken together, these key points position FGC as a promising addition to the current interpretability landscape, complementing existing methods with structure-aware, model-informed insights.

The AML case study further illustrates the value of FGC in deriving meaningful biological insights from complex models. Although trained on a binary AML classification task, the model internally distinguished between multiple leukemia types, including CML, CLL, and ALL, and revealed confounder-driven variation linked to study origin. By enabling a structured analysis of gene expression patterns, FGC helped to disentangle disease-related signals from those driven by cell-of-origin effects. This distinction is critical in biomedical applications: while immune response related genes emerged as predictive, further investigation revealed that their signal likely reflected immune cell composition rather than leukemia-specific dysregulation. These findings highlight the importance of combining explainability tools with expert knowledge when interpreting XAI outputs. In clinical settings, models that unintentionally depend on indirect signals, such as cell-of-origin effects, may achieve high accuracy yet yield misleading conclusions or harmful prediction if deployed unchecked. FGC helps to mitigate such risks by exposing the internal logic of the model, enabling careful examination of the subgroups and features that guide predictions. In translational research and diagnostics, such interpretability is not optional: it is essential for distinguishing robust biomarkers from confounders, guiding patient stratification, and building trust in model-informed decision-making. These results emphasize the critical role of explainability in safe and responsible AI deployment.

By revealing both informative structure and potential biases, it allows users to understand not only how well a model performs, but also why it performs well and whether the underlying patterns are meaningful. In doing so, FGC supports both scientific discovery and responsible AI deployment, especially in domains where fairness, robustness, and trust are essential.

Looking ahead, while FGC is already compatible with RF regressors, its application to regression tasks poses unique challenges. Trees of RF regressors are typically deeper and more branched than those of classifiers, which increases the sparsity of shared decision paths and makes it more difficult to derive meaningful clusters from the resulting proximity matrix. Future work may address this limitation by incorporating internal tree pruning strategies during the FGC proximity computation to improve clustering robustness. Additional extensions include adapting FGC to support other tree-based models such as gradient-boosted decision trees, and integrating alternative clustering algorithms beyond k-medoids to enhance flexibility and generalizability across diverse modeling scenarios.

Ultimately, FGC bridges the gap between model performance and interpretability and provides a framework for data-driven hypothesis generation, model auditing, and the transparent integration of AI systems in real-world applications.

\section{Methods}\label{sec4}

\subsection*{Distance Matrix Computation}\label{sec4_1}

The first step in the FGC algorithm is the computation of a model-informed distance matrix that captures the dissimilarities between input data points according to the RF tree traversal patterns. Compared to standard distance metrics like Euclidean or Manhattan distance, which rely only on the feature space, this distance matrix leverages the RF’s internal proximity structure, providing a learned representation of instance similarity. The proximity matrix $M^{\text{proximity}} \in \mathbb{R}^{n \times n}$ is a square matrix, where $n$ is the number of input data points. Each entry $M_{ij}^{\text{proximity}}$ represents how frequently two data points $i$ and $j$ end up in the same terminal nodes across all trees in the RF. It is defined as:
\begin{equation}
M_{ij}^{\text{proximity}} = \frac{m_{ij}}{N}
\end{equation}
where $m_{ij}$ is the number of trees in which $i$ and $j$ share the same terminal node, and $N$ is the total number of trees in the RF model. The obtained proximity matrix is symmetric, with values ranging from 0 (never in the same leaf) to 1 (always in the same leaf). While the proximity matrix quantifies similarity, clustering algorithms typically require dissimilarity (distance) as input. Therefore, the proximity values are transformed into distance values using:
\begin{equation}
M_{ij}^{\text{distance}} = 1 - M_{ij}^{\text{proximity}}
\label{eq:dist_mat}
\end{equation}
According to Breiman \citep{breiman2008manual}, these distances behave like squared distances in a Euclidean space, making them suitable for many clustering algorithms that can help to segment the input space.

We provide a detailed runtime and memory benchmarking analysis of the implemented distance matrix computation, including its scaling behavior with increasing sample sizes, in Appendix~\ref{secA3}.

\subsection*{Clustering Optimization}

To uncover coherent subgroups of samples that follow similar decision pathways in the RF, we perform clustering on the RF-derived distance matrix $M_{ij}^{\text{distance}}$ as defined in Equation ~\ref{eq:dist_mat}. 

\paragraph{Clustering Algorithms}\label{sec4_2}

Since the RF-derived distance matrix reflects pairwise distances rather than coordinates in a metric space, we use $k$-medoids clustering, which can operate directly on arbitrary distances by selecting actual data points as cluster centers (medoids) and minimizing the sum of distances between the data points and their nearest medoids. Because the exact solution to $k$-medoids clustering is NP-hard, the Partitioning Around Medoids (PAM) algorithm is commonly used as an efficient approximation. As shown in Algorithm~\ref{alg:pam}, PAM heuristically searches for a near-optimal set of medoids by iteratively swapping medoids with non-medoids to reduce clustering costs. The clustering cost is quantified using the internal clustering evaluation metric (\textit{inertia}) that captures cluster compactness by summing the distances between each data point and its assigned medoid. Specifically, the inertia $I_k$ for a clustering into $k$ clusters is defined as:
\begin{equation}
I_k = \sum_{j=1}^k \sum_{x_i \in C_j} d(x_i, m_j),
\end{equation}
where $d(x_i, m_j)$ denotes the pairwise distance between data point $x_i$ and its assigned medoid $m_j$, and $C_j$ represents the set of points in cluster $j$. Lower values of $I_k$ indicate more compact and coherent clusters.
\begin{algorithm}
\caption{PAM (Partitioning Around Medoids)}\label{alg:pam}
\begin{algorithmic}[1]
\Require Distance matrix $D$, number of clusters $k$
\Ensure Cluster assignments and medoids minimizing total inertia
\State \textbf{Initialize:} Select $k$ initial medoids using a greedy heuristic.
\Repeat
    \State \textbf{Assign:} Assign each data point to the nearest medoid using $D$.
    \State \textbf{Compute Inertia:} $I_k = \sum_{j=1}^k \sum_{x_i \in C_j} d(x_i, m_j)$
    \ForAll{medoid $m$}
        \ForAll{non-medoid point $o$}
            \State Swap $m$ and $o$
            \State Compute new inertia $I_k'$
            \If{$I_k' < I_k$}
                \State Accept the swap
                \State $I_k \gets I_k'$
            \EndIf
        \EndFor
    \EndFor
\Until{No beneficial swaps or maximum iterations reached}
\State \Return Clustering with medoids
\end{algorithmic}
\end{algorithm}\\
For large datasets, the quadratic growth of the distance matrix and clustering operations in $O(n^2k)$ make standard $k$-medoids clustering computationally infeasible. To address this limitation, we implemented a modified variant of PAM known as CLARA (Clustering Large Applications) \cite{kaufman2009finding}, which estimates medoids by performing clustering on multiple random subsamples and subsequently assigns the entire dataset to clusters based on the identified representative medoids. The full algorithm is outlined in Algorithm ~\ref{alg:clara}.

We provide a benchmark of runtime and memory usage for the available PAM algorithms, as well as an evaluation of the scalability of our CLARA implementation on large datasets, in Appendix~\ref{secA3}.

\begin{algorithm}

\caption{CLARA (Clustering Large Applications)}\label{alg:clara}
\begin{algorithmic}[1]
\Require RF model $rf$, number of clusters $k$, number of iterations $T$, subsample size $n$
\Ensure Cluster assignments and medoids minimizing total inertia
\For{each iteration $t = 1$ to $T$}
    \State Randomly sample a subset of size $n$
    \State Compute the distance matrix on the subset using $rf$-derived proximities
    \State Apply $k$-medoids to the subset to identify the medoids
    \State Assign all remaining data points in the full dataset to their nearest medoid
    \State Compute inertia for the full dataset clustering
\EndFor
\State \Return Clustering with the lowest total inertia across all $T$ iterations
\end{algorithmic}
\end{algorithm}

\paragraph{Selection of the Number of Clusters ($k$)} 

Determining the optimal number of clusters is crucial for obtaining a meaningful segmentation of the input sample space. To this end, we introduce a scoring framework that selects $k$ by jointly minimizing model bias and constraining model variance. The model bias assesses the ability of the clustering to approximate the target distribution, while the model variance penalizes unstable solutions.

For \textbf{classification tasks}, we define model bias using a class-balanced Gini impurity score:
\begin{equation}
\text{Bias}_k = \sum_{i=1}^{k} \left( 1 - \sum_{g=1}^{G} b_{i,g}^2 \right),
\end{equation}
where $b_{i,g} = \frac{1}{\sum_{g=1}^{G} \frac{p_{i,g}}{q_g}} \times \frac{p_{i,g}}{q_g}$ is the class-balanced frequency of class $g$ in cluster $i$, $p_{i,g}$ is the proportion of samples of class $g$ in cluster $i$, and $q_g$ is the overall proportion of class $g$ in the dataset. $b_{i,g}$ re-weights $p_{i,g}$ by the inverse of $q_g$, ensuring that minority classes contribute equally to the impurity calculation. This is crucial in imbalanced classification problems, where otherwise large classes could dominate the impurity score.

For \textbf{regression tasks}, we assess the model bias by computing the normalized within-cluster variance:
\begin{equation}
\text{Bias}_k = \frac{\sum_{j=1}^{k} n_j \cdot \text{Var}(y_j)}{n \cdot \text{Var}(y)},
\end{equation}
where $n_j$ is the number of samples in cluster $j$, $\text{Var}(y_j)$ is the variance of the target values within that cluster, $n$ is the total number of data points, and $\text{Var}(y)$ is the variance of the target values over the entire dataset. In both cases, lower values indicate that the clustering aligns well with the target variable, either through high class purity within clusters or by showing low within-cluster variation.

Model variance is assessed through clustering \textit{stability}, measured via the Jaccard similarity between the original clustering $A$ and bootstrapped clustering $B_b$:
\begin{equation}
\text{Stability}_i = \frac{1}{n} \sum_{b=1}^{n} \frac{|A \cap B_b|}{|A \cup B_b|}.
\end{equation}
Only clusterings with average stability over $n$ bootstrapped clusterings above 0.6 are retained because this threshold is  considered indicative of clusterings that are robust to resampling and likely to generalize \cite{hennig2007cluster}. The optimal value of $k$ is chosen as the solution with minimum bias among the stable candidates. 

\subsection*{Feature Importance}\label{sec4_3}

The FGC algorithm estimates local feature importance by quantifying how much a feature's distribution within a given cluster deviates from its distribution across the entire dataset. For each feature and cluster, this is achieved by comparing the within-cluster distribution to the global distribution. Two distance measures are available to quantify these deviations: the Wasserstein distance and the Jensen-Shannon distance. 

To derive global feature importance, the algorithm computes the average of the local importances across all clusters. This aggregation captures the overall relevance of each feature across the clustering structure and enhances interpretability. A key challenge in comparing distributions lies in handling differences between numeric and categorical features while ensuring comparability. 

The Wasserstein distance, also known as Earth Mover's or Kantorovich-Rubinstein distance in the one-dimensional case, is naturally suited for continuous data. It represents the minimal effort required to transform one distribution into the other. On the real line, the Wasserstein distance is defined as 
\begin{equation}
W_1 (\mu, \nu) = \int_{-\infty}^{\infty} |F(x) - G(x)| dx, 
\end{equation}
where $F$ and $G$ are the cumulative distribution functions of the probability measures $\mu$ and $\nu$, respectively \citep{wasserstein_1, wasserstein_2}. To apply the Wasserstein distance to categorical features, we one-hot encode each categorical feature. The distance is then computed separately for each resulting binary variable. The maximum distance across the categories is used to quantify the importance of the feature. This approach captures the most pronounced distributional shift.

The Jensen-Shannon distance, by contrast, is well-suited for comparing two discrete probability distributions. It is defined as the square root of the Jensen-Shannon divergence, which is a symmetric version of the Kullback–Leibler divergence: 
\begin{equation}
KL(p:q) = \sum_{i=0}^{k} p_i \log \frac{p_i}{q_i},
\end{equation}
where $p=(p_0, \ldots, p_k)$ and $q=(q_0, \ldots, q_k)$ are probability mass functions. While the Kullback-Leibler divergence measures how well the distribution $q$ approximates $p$, it is not symmetric. To address this, the Jensen-Shannon divergence is defined as: 
\begin{equation}
JSD(p, q) = \frac{1}{2} \sum_{i=0}^{k} \left( p_i \log \frac{2p_i}{p_i + q_i} + q_i \log \frac{2q_i}{p_i + q_i} \right), 
\end{equation}
where the average distribution $\frac{p+q}{2}$ is a categorical distribution serving as a common, symmetric reference distribution. \citep{jensen_shannon_1} To apply the Jensen-Shannon divergence to numeric features, we discretize them by binning the continuous values into intervals, thereby transforming them into categorical distributions. First, the number of bins is determined using the Freedman–Diaconis rule, with bounds to ensure a reasonable number of bins. Then, quantile-based binning is applied, creating approximately equally populated bins of the overall data. This strategy ensures robustness even in the presence of skewed or heavy-tailed distributions.

In practice, the Jensen-Shannon distance offers greater robustness for discrete features, while the Wasserstein distance provides more interpretable values for continuous ones. Regardless of the distance metric used, local importance scores are normalized within each cluster by dividing all feature distances by the maximum value observed in that cluster. This ensures that the most divergent feature is assigned a normalized importance of 1.0, with all others scaled proportionally. This normalization enables comparability of importance values across clusters while preserving the relative ranking of features within each cluster.

\subsection*{Benchmark Study}\label{sec4_4}

\paragraph{Simulated Data}

We generated a synthetic dataset consisting of 600 observations assigned to two primary classes. Only class 1 was subdivided into latent subclasses, while class 2 remained homogeneous. The dataset included five features: three informative features that encode subclass structure within class 1, and two noise features that are unrelated to either class or subclass labels. Within class 1, we defined three subclasses. Two of these (100 samples each) were separated by Features 1 and 2, which were drawn from normal distributions with subclass-specific means. Feature 3, also normally distributed, distinguished a third subclass of 200 samples. Together, these three features introduced a four-subclass structure nested within the broader binary class assignment. To test model behavior in the presence of irrelevant inputs, we added two noise features: one continuous variable drawn from a standard normal distribution, and one binary variable, both independent of class and subclass labels. The overall feature distribution is shown in Figure~\ref{fig:simulated_data}.

\paragraph{Clustering Approaches}

Several approaches exist for clustering high-dimensional data. A common baseline is k-medoids clustering, which partitions observations based on pairwise distances such as the Euclidean distance \citep{rdusseeun1987clustering}. Each cluster is represented by its medoid, the most centrally located data point, and clustering aims to minimize the total distance between points and their assigned medoid.

An alternative strategy is the unsupervised RF clustering approach described by Shi and Horvath \citep{shi2006unsupervised}. This method constructs a proximity matrix by training a Random Forest classifier to distinguish real observations from synthetic noise, where the latter is sampled either from the product of empirical marginal distributions or uniformly from the observed feature ranges. The proximity scores derived from the trained RF model are then used to perform k-medoids clustering.

Both k-medoids and unsupervised RF clustering operate without access to class labels and are therefore purely unsupervised. In contrast, FGC incorporates class label information directly into the clustering process. This supervision allows FGC to recover label-relevant subgroups that may remain hidden in unsupervised settings. Furthermore, FGC provides cluster-specific feature importance scores, enabling the identification of features most relevant to each subgroup.

\paragraph{Feature Importance Approaches}

Several general-purpose explainable AI methods are commonly used to assess feature importance, including permutation-based measures such as MDA, and model-agnostic approaches such as SHAP and LIME. For tree-based models like Random Forests, treeSHAP provides an efficient, model-specific implementation of SHAP values. In addition, traditional internal metrics such as MDI offer alternative strategies for estimating the influence of individual features on model predictions.

Mean Decrease in Accuracy (MDA) is a permutation-based method for estimating feature importance. It assesses how much the model’s predictive accuracy decreases when the values of a single feature are randomly permuted, thereby breaking its association with the target. A large drop in accuracy indicates that the feature is important for the model’s predictions \citep{breiman2001random}. However, MDA can underestimate the importance of correlated features, as permuting one may not significantly impact accuracy if the model can still rely on correlated alternatives.

Mean Decrease in Impurity (MDI) is a feature importance measure specific to tree-based models. It quantifies how much each feature reduces impurity (e.g., Gini or entropy) across all splits in a Random Forest, averaged over all trees \citep{breiman2001random}. In contrast to MDA, MDI does not require permuting input features and is computed directly from the training process, but it may introduce bias toward features with more categories or continuous values. 

SHapley Additive exPlanations (SHAP) is a model-agnostic method that quantifies the contribution of each feature to a specific prediction. It is based on cooperative game theory and computes the marginal contribution of a feature by comparing the model’s output with and without the feature. The final SHAP value is obtained by averaging these marginal contributions across all possible feature subsets, weighted by their size. For tree-based models, TreeSHAP offers an efficient algorithm for computing SHAP values for ensemble tree-based methods \citep{lundberg2017unified}.

Local Interpretable Model-agnostic Explanations (LIME) provides local interpretability by approximating a complex model in the vicinity of a specific instance. It does so by generating perturbed versions of the input sample, making predictions with the original model, and fitting a simple, interpretable surrogate model (e.g., a linear regressor) to this locally weighted dataset. The coefficients of the surrogate model indicate the local importance of each feature, offering insight into the model’s behavior around the instance of interest \citep{ribeiro2016should}.

\paragraph{Method Comparison}

We trained a k-medoids clustering model on the simulated dataset using the \texttt{fasterpam} algorithm from the \textit{kmedoids} Python package, specifying four medoids and a maximum of 200 iterations \citep{schubert2022fast}. For unsupervised Random Forest (RF) clustering, we used the \textit{randomForest} \texttt{R} package with 2000 trees, and performed clustering using the \textit{cluster} \texttt{R} package \citep{liaw2002classification, maechler2025cluster}. To enable comparison across explainability methods, we also implemented a Random Forest classifier in Python using \textit{scikit-learn}, from which both MDI and MDA were computed \citep{scikitlearn}. This model was trained with five-fold cross-validation over a predefined hyperparameter grid: \texttt{max\_depth: [2, 3]}, \texttt{max\_features: [sqrt, log2]}, \texttt{bootstrap: [True]}, \texttt{max\_samples: [0.8]}. For MDA, each feature was permuted 50 times using $80\%$ of the samples. For FGC, clustering was performed using \texttt{fasterpam} with a maximum of 200 iterations, and feature relevance was quantified using the Wasserstein metric, suitable for our mainly continuous features. Local interpretable explanations were generated using the \textit{lime} Python package and feature importances were aggregated by averaging weights across all individual predictions \citep{ribeiro2016should}. SHAP values were computed using texttt{treeSHAP} via the textit{shap} Python package \citep{lundberg2020local2global}. Both LIME and SHAP analyses were done directly on the training set. The notebooks for the data simulation and method comparison of the benchmark study are provided at: \url{https://github.com/HelmholtzAI-Consultants-Munich/fg-clustering/tree/main/publication/benchmark}.

\subsection*{AML Dataset}\label{sec4_5}

We used a publicly available RNA-seq dataset from Warnat-Herresthal et al. \citep{warnat2020scalable} to model and explain Acute Myeloid Leukemia (AML) status using gene expression data. The dataset contains log$_2$-transformed, normalized gene expression values for over 12{,}000 genes across a heterogeneous cohort of individuals, including AML patients, patients with other leukemia types (e.g., CLL, CML, ALL, MDS), healthy controls, and individuals with non-leukemic diseases. Gene expression profiling was performed using both microarrays and RNA-seq. For our case study, we focus exclusively on the RNA-seq subset due to its higher gene expression resolution. The processed data is distributed as part of a Docker image released by the original authors. We extracted the RNA-seq dataset from the image, converted the relevant objects to CSV using \texttt{R}, and merged gene expression values with associated sample metadata. No additional normalization or transformation was applied to the obtained data. Minor spelling corrections were made to metadata entries for tissue and disease categories. 

Prior to model training, we conducted an exploratory analysis of the dataset to assess sources of variability that may impact downstream modeling (Figure ~\ref{fig:AML_case_study_dataset}). Samples were drawn from over 20 independent studies and two primary tissue types, i.e. peripheral blood mononuclear cells (PBMCs) and bone marrow (BM), introducing potential batch effects and tissue-specific confounding. Dimensionality reduction using PCA and UMAP revealed that sample organization is strongly influenced by study origin, sample tissue source, and disease label, highlighting the need of mitigating technical confounders when training and interpreting the model. To address these confounding sources of variation, we trained the RF classifier that generalizes across studies. Therefore, we used a stratified group $k$-fold cross-validation strategy, with the study identifier (GEO accession number) as the grouping variable as suggested in the original publication by Warnat-Herresthal et al. \citep{warnat2020scalable}. This approach ensures that samples from the same study are not split between training and test folds, thereby preventing data leakage and enabling model evaluation on entirely unseen cohorts. Stratification preserved class balance across folds. The training set comprised $66\%$ of the data (783 samples), including 245 AML (cancer) and 538 non-AML (control) samples from 18 studies. The remaining $34\%$ (398 samples), including 263 AML (cancer) and 135 non-AML (control) samples from 5 previously unseen studies, was held out for testing. Hyperparameter tuning was conducted via nested cross-validation using the same stratified group folds, optimizing over a predefined parameter grid: \texttt{max\_depth: [5, 10, 20]}, \texttt{max\_features: [sqrt, log2]}, \texttt{bootstrap: [True]}, \texttt{max\_samples: [0.8]}, \texttt{n\_estimators: [1000]}, \texttt{criterion: [gini]}, \texttt{min\_samples\_leaf: [1, 2, 5]}. This setup accounts for potential study-specific batch effects we observed in Supplementary Figure ~\ref{fig:AML_case_study_dataset}. The final model achieved $96.6\%$ balanced accuracy on an independent test set, demonstrating strong predictive performance and robustness to study-level heterogeneity. 

After final model selection, we applied FGC to the trained RF model to extract interpretable cluster structures based on shared decision paths. Clustering was performed using the k-medoids algorithm over a candidate range of \texttt{k=(2,10)}, with parameters \texttt{method="fasterpam"}, \texttt{init="random"}, and \texttt{max\_iter=100}. To ensure robustness, clustering stability was assessed using the Jaccard index with 100 bootstrap iterations and a sample size of $80\%$, discarding unstable configurations below a threshold of 0.9 using the parameters: \texttt{JI\_bootstrap\_iter=100}, \texttt{JI\_bootstrap\_sample\_size=0.8}, \texttt{JI\_discart\_value=0.9}. The final clustering yielded eight stable clusters, each representing a subgroup of samples that follow similar decision paths through the forest (Table~\ref{tab:FGC_cluster_stability}). For each cluster, we computed local feature importance scores to identify genes that contributed most strongly to the model’s decisions. Most clusters were defined by distinct gene expression signatures among their top 30 features, though three genes (BTN3A3, CD48, GVINP1) appeared in both AML-enriched and control-associated clusters (Figure~\ref{fig:AML_case_study_feature_importance}). To further interpret the model’s decision logic and assess potential confounders, we visualized decision paths using a targeted gene panel composed of the top 30 genes from the AML-enriched cluster, along with genes known to be overexpressed in AML (BAALC, FLT3, HOXA9, MECOM, WT1) \citep{handschuh2019not}. Metadata annotations, including sample tissue type, study origin, and disease label, were incorporated to assess the influence of technical or biological confounders. The full analysis pipeline, including exploratory data analysis and FGC interpretation, is available at \url{https://github.com/HelmholtzAI-Consultants-Munich/fg-clustering/tree/main/publication/aml_case_study}.

\backmatter

\bmhead{Data availability}
The acute myeloid leukemia (AML) dataset used in this study was previously published by Warnat-Herresthal et al. \citep{warnat2020scalable} and is publicly available via Docker Hub at: \url{https://hub.docker.com/r/schultzelab/aml_classifier}.\\

\bmhead{Code availability}
The implementation of the Forest Guided Clustering algorithm is available as an open-source Python package at:
\url{https://github.com/HelmholtzAI-Consultants-Munich/fg-clustering}.
The repository includes the full source code, installation instructions, and multiple tutorials that demonstrate the application of FGC to different datasets. These tutorials also provide guidance on scaling the algorithm to large datasets. The code is released under the MIT License and is compatible with standard machine learning frameworks.

\bmhead{Acknowledgements}
We thank the Helmholtz AI Consultants Team for Health for their continuous feedback and support throughout the development of this work. We also acknowledge the broader Helmholtz AI community and the consultant teams for Energy, Earth and Environment, Matter, and Information for their inspiring discussions and valuable input. Special thanks go to Sara Hetzel for her insightful discussions on the AML case study and for providing feedback on the corresponding section of the manuscript.

\bmhead{Author information}
L.B.A.S. led the conceptual development of the FGC method, developed the core algorithm, implemented the main functionality of the Python package, and authored the accompanying tutorials and documentation. She applied FGC to the AML dataset, conducted the runtime and memory profiling, and was the primary author of the manuscript. G.M. and R.L.G. conceptualized and implemented the feature importance framework within FGC, did the benchmark comparison to other xai methods and contributed to writing the results (benchmark) and methods sections (feature importance) of the manuscript. D.T. contributed to the early development of the FGC Python package and co-developed the initial software architecture. H.P. contributed to the implementation of FGC for large-scale datasets and co-authored the user tutorials. M.P. supervised the project and revised the manuscript. All authors contributed to the final version of the manuscript and discussed the manuscript.\\

\bmhead{Corresponding Authors}
Correspondence to: \href{mailto:lisa.barros@helmholtz-munich.de}{Lisa Barros de Andrade e Sousa}, \href{mailto:marie.piraud@helmholtz-munich.de}{Marie Piraud}

\clearpage
\begin{appendices}

\section{Extended Results: Benchmark Study}\label{secA1}

To illustrate the differences between Forest-Guided Clustering (FGC) and unsupervised clustering approaches or traditional explainability methods, we conducted a comparative analysis on a simulated dataset composed of two main classes, one of which contains three latent subclasses.

\begin{figure}[ht]
    \centering
    \includegraphics[width=\textwidth]{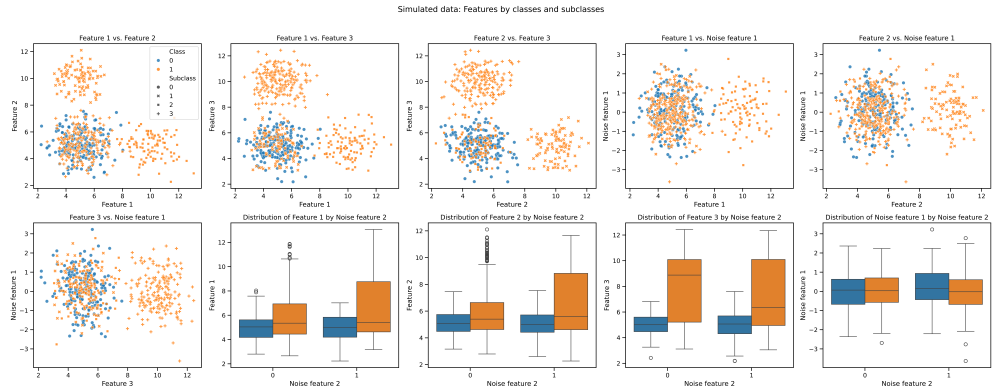}
    \caption{\textbf{Structure of the simulated dataset.}
    The simulated dataset consists of two main classes, where one class is subdivided into three latent subclasses. Three informative features encode this subclass structure, while two additional features (one continuous, one binary) serve as noise and are uncorrelated with the class labels.}
    \label{fig:simulated_data}
\end{figure}

\clearpage
\section{Extended Results: AML Case Study}\label{secA2}

To demonstrate the insights FGC can offer for high-dimensional biomedical data, we applied it to a curated RNA-seq dataset focused on Acute Myeloid Leukemia (AML), originally published by Warnat-Herresthal et al. \citep{warnat2020scalable}. This dataset contains log$_2$-transformed, normalized gene expression profiles across more than 12,000 genes, covering a heterogeneous cohort of AML patients, individuals with other leukemia subtypes, non-leukemic conditions, and healthy controls.

We sought to evaluate the biological relevance of the most important model-derived features identified by FGC. Specifically, we analyzed gene expression patterns across FGC-defined clusters to determine whether they aligned with known biological pathways, reflected immune lineage composition, or suggested disease-related dysregulation. This analysis helps to distinguish whether the model recapitulates established AML signatures, captures novel biology, or reflects potential confounding variation.

\paragraph{Immune Response and Cytokine Signaling}
Several top-ranked genes, including BTN3A1, BTN3A2, BTN3A3, CD48, IL16, and GVINP1, are known immune markers. Their expression patterns, particularly the elevated levels in PBMC-rich control clusters (Clusters 3 and 5), suggest they may primarily reflect immune cell composition rather than leukemia-driven dysregulation. This aligns with their known biology as immune markers expressed in T cells, monocytes, and other lymphoid lineages. Reduced expression in AML, CML, and MDS clusters is consistent with a loss of mature immune cells. While CD48 has been reported to be downregulated in AML \citep{wang2020acute, elias2014immune}, other genes in this group, including IL16 and GVINP1, are not established AML drivers. Hence, these profiles are more plausibly explained by cell-of-origin differences than by disease-specific dysregulation. 

\paragraph{Chromatin and Epigenetic Regulators}
Unlike immune-related genes, the expression patterns of chromatin-associated genes such as MORF4L2, H2AFY, CHD1L, and SETMAR more plausibly reflect mechanisms linked to malignant transformation. MORF4L2, although not previously linked to AML or other cancer types, showed pronounced upregulation in the AML-enriched cluster, suggesting a potential but yet uncharacterized role in leukemic gene regulation. H2AFY, a histone variant implicated in transcriptional repression, has been shown to critically influence leukemia stem cell function and differentiation in AML, and was consistently elevated in myeloid malignancies (AML, CML, MDS) while remaining low in healthy and lymphoid clusters \citep{hsu2022functional, hsu2019functional}. CHD1L, a chromatin remodeler broadly recognized as an oncogene in solid tumors \citep{soltan2023pan}, also showed its highest expression in AML, although its specific role in hematologic cancers remains less defined. Finally, SETMAR, a fusion gene involved in DNA repair and chromatin organization, has been found upregulated in AML bone marrow samples, and several transcript variants have been associated with tumorgenisis in leukemia \citep{sirma2004set, handschuh2018gene, cristobal2012overexpression}. 

\paragraph{Transcriptional Regulators}
We next examined transcriptional regulators, including ZNF629, BZW2, HMGXB3, and SCML2. These genes showed distinct expression patterns across clusters, potentially reflecting malignancy-related processes. ZNF629, though not previously linked to AML, showed its highest expression in the AML-enriched cluster. BZW2, implicated in AML stem cell survival, was elevated in AML, CML, and MDS \citep{nachmias2022ipo11}. HMGXB3, part of a gene family associated with myeloid malignancies, showed increased expression in leukemic clusters but low levels in controls \citep{minervini2020hmga, lin2021roles}. SCML2, previously reported as overexpressed in specific AML subtypes, was most elevated in Cluster 1 and less so across other leukemias, possibly reflecting a transcriptional repression program associated with leukemic transformation \citep{grubach2008gene}. 

\paragraph{Catalytic Activity}
Another category, including catalytic enzymes such as IPCEF1, ICK, PLCL2, CEPT1, UCK2, NUDT15, OCRL, ACOT11, CCNB1IP1, and BMP1, showed expression patterns that may reflect a mix of malignancy-associated processes and cell-of-origin effects. For instance, IPCEF1, ICK, PLCL2, and CEPT1 showed higher expression in control clusters or lymphoid leukemias, consistent with their roles in immune cell function. In contrast, UCK2, NUDT15, ACOT11, CCNB1IP1, and BMP1 exhibited elevated expression in AML or other myeloid malignancies. These findings are in line with previous studies reporting frequent upregulation of UCK2 in diverse cancer types, leukemia-suppressive roles of NUDT15 in murine models, and associations of ACOT11 and CCNB1IP1 expression with poor prognosis in AML \citep{tian2025uridine, wu2019uridine, wang2024nudt15, luo2016microarray, ersvaer2007cyclin, manoochehrabadi2024upregulation}. BMP1, similarly, aligns with broader involvement of BMP signaling in AML progression \citep{zylbersztejn2018bmp}. The remaining gene, OCRL, showed moderate AML-specific elevation without prior AML linkage, suggesting a potential but unexplored role. 

\paragraph{Others}
Lastly, we examined the remaining model-derived features with functionally diverse roles, AKTIP, RETREG3, VDAC2, TWSG1, KCTD3, and ANKRD28, to assess whether their expression patterns reflect malignancy-associated dysregulation or lineage-related differences. Several genes, such as AKTIP and RETREG3, were more highly expressed in control clusters and lymphoid malignancies, which may suggest a role linked to cell physiology rather than malignant transformation. In contrast, genes such as VDAC2, TWSG1, KCTD3, and ANKRD28 exhibited elevated expression in AML-enriched clusters, with relatively low expression in control or non-myeloid malignancies. This pattern could point to a potential involvement in AML-associated pathways, particularly for ANKRD28, which has been implicated in poor prognosis in AML \citep{de2011gene}. TWSG1, VDAC2 and KCTD3 also demonstrated AML-biased upregulation without broad elevation across other lineages, possibly reflecting roles in AML-specific signaling or transformation. \\\\
An overview of the top model-derived genes, including their functional annotations and interpretations regarding potential cell-of-origin versus disease-specific expression patterns, is provided in Table~\ref{tab:gene_level_summary}.

\begin{figure}[ht]
    \centering
    \includegraphics[width=\textwidth]{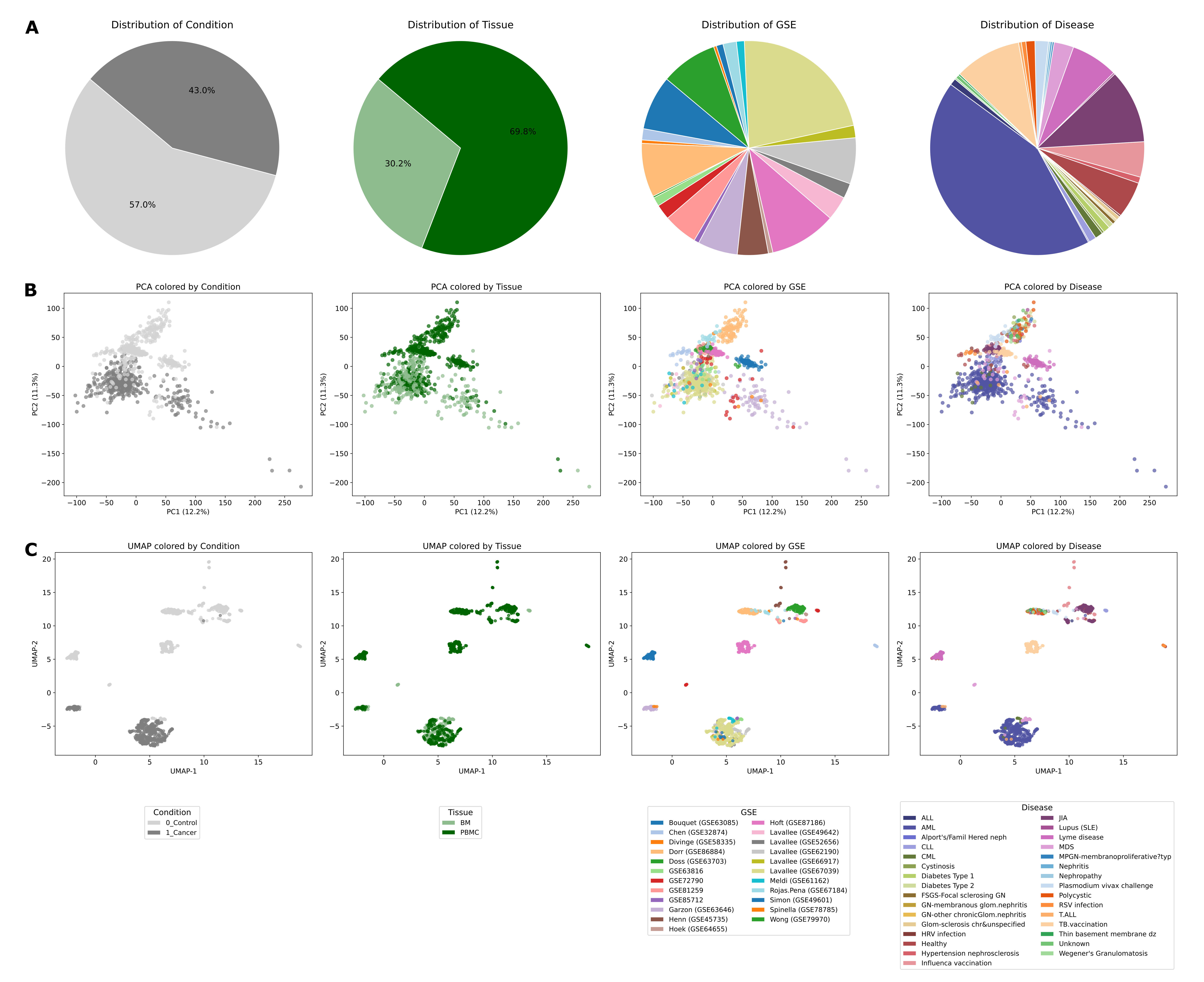}
    \caption{\textbf{Exploratory analysis of the RNA-Seq dataset.}
    (A) Pie charts summarizing dataset composition with respect to condition, tissue source, study origin (GSE accession), and disease annotation. 
    (B) PCA and 
    (C) UMAP colored by different metadata types reveal structure in the data. Both dimensionality reduction techniques show clustering patterns associated with condition, tissue type, study origin, and disease labels, highlighting potential batch effects and confounders that should be considered in model development and interpretation.}
    \label{fig:AML_case_study_dataset}
\end{figure}

\begin{table}[ht]
\centering
\caption{\textbf{Summary of FGC Performance.} For each $k$, the table reports the model score, whether the clustering passed the stability threshold (\texttt{Stable}), the mean Jaccard index across clusters, and the per-cluster Jaccard indices.}
\label{tab:FGC_cluster_stability}
\small
\begin{tabular}{|c|c|c|c|p{7.5cm}|}
\hline
\textbf{k} & \textbf{Score} & \textbf{Stable} & \textbf{Mean JI} & \textbf{Cluster-wise JI} \\
\hline
2  & 0.0073 & \checkmark  & 1.00 & \{1: 1.00, 2: 1.0\} \\
3  & 0.0048 & \checkmark  & 0.99 & \{1: 0.98, 2: 0.99, 3: 1.0\} \\
4  & 0.0027 & \checkmark  & 0.94 & \{1: 0.99, 2: 0.98, 3: 0.96, 4: 0.81\} \\
5  &        &             & 0.89 & \{1: 0.99, 2: 0.98, 3: 0.87, 4: 0.72, 5: 0.91\} \\
6  & 0.00062 & \checkmark & 0.96 & \{1: 0.92, 2: 0.99, 3: 0.99, 4: 0.97, 5: 0.87, 6: 1.0\} \\
7  & 0.00053 & \checkmark & 0.95 & \{1: 0.93, 2: 0.99, 3: 0.88, 4: 0.99, 5: 0.85, 6: 0.99, 7: 1.00\} \\
8  & 0.00046 & \checkmark & 0.94 & \{1: 0.75, 2: 0.99, 3: 0.97, 4: 0.99, 5: 0.97, 6: 0.89, 7: 1.00, 8: 0.94\} \\
9  &         &            & 0.87 & \{1: 0.99, 2: 0.99, 3: 0.44, 4: 0.98, 5: 0.88, 6: 0.92, 7: 1.00, 8: 0.90, 9: 0.69\} \\
10 &         &            & 0.85 & \{1: 0.83, 2: 0.99, 3: 0.66, 4: 0.98, 5: 0.45, 6: 0.89, 7: 1.00, 8: 0.77, 9: 0.99, 10: 0.93\} \\
\hline
\end{tabular}
\end{table}

\begin{figure}[ht]
    \centering
    \includegraphics[width=\textwidth]{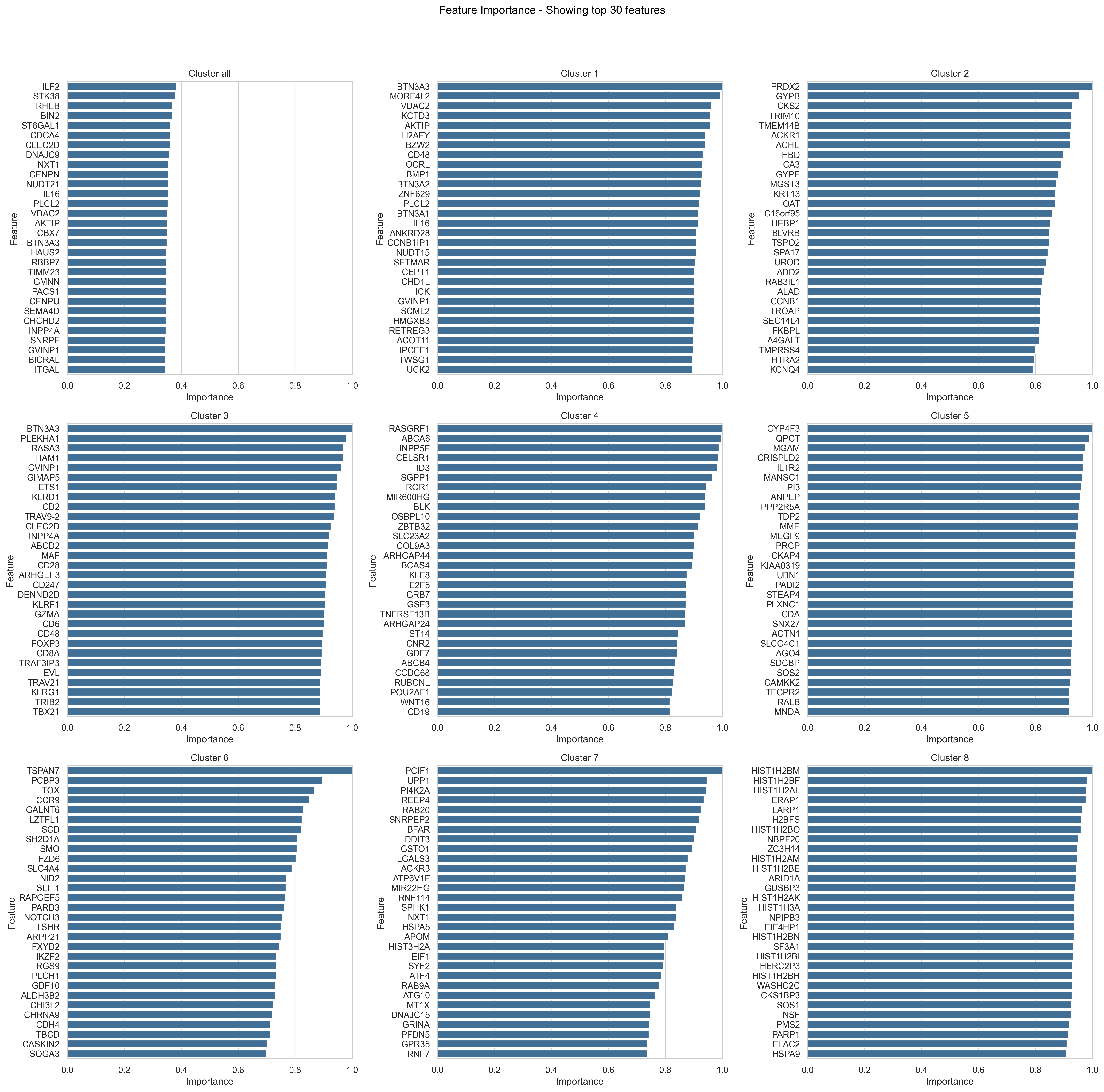}
    \caption{\textbf{Global and local feature importance for the AML case study.}
    FGC revealed eight stable decision clusters within the RF model. For each cluster, local feature importance was computed based on the Wasserstein distance, ranking genes by their distinctiveness relative to a background distribution. The top 30 most important genes per cluster are shown, as well as the top 30 globally importance genes computed by averaging local scores across clusters. While most clusters exhibit unique gene signatures, BTN3A3, CD48, and GVINP1 are shared between the AML-associated cluster (Cluster 1) and the healthy control cluster (Cluster 3).}
    \label{fig:AML_case_study_feature_importance}
\end{figure}

\begin{table}[htbp]
\centering
\scriptsize
\begin{tabular}{lcccccccc}
\toprule
\textbf{Disease} & \textbf{1} & \textbf{2} & \textbf{3} & \textbf{4} & \textbf{5} & \textbf{6} & \textbf{7} & \textbf{8} \\
\midrule
ALL & 1 & 0 & 0 & 1 & 0 & 10 & 0 & 0 \\
AML & 245 & 0 & 0 & 0 & 0 & 0 & 0 & 0 \\
Alport's/Famil Hered neph & 0 & 0 & 1 & 0 & 1 & 0 & 0 & 0 \\
CLL & 0 & 0 & 0 & 13 & 0 & 0 & 0 & 0 \\
CML & 0 & 6 & 0 & 0 & 0 & 0 & 8 & 0 \\
Cystinosis & 0 & 0 & 3 & 0 & 1 & 0 & 0 & 0 \\
Diabetes Type 1 & 0 & 0 & 8 & 0 & 4 & 0 & 0 & 0 \\
Diabetes Type 2 & 0 & 0 & 6 & 0 & 3 & 0 & 0 & 0 \\
FSGS-Focal sclerosing GN & 0 & 0 & 4 & 0 & 2 & 0 & 0 & 0 \\
GN-membranous glom.nephritis & 0 & 0 & 1 & 0 & 1 & 0 & 0 & 0 \\
GN-other chronicGlom.nephritis & 0 & 0 & 2 & 0 & 0 & 0 & 0 & 0 \\
Glom-sclerosis chr\&unspecified & 0 & 0 & 4 & 0 & 3 & 0 & 0 & 0 \\
HRV infection & 0 & 0 & 3 & 0 & 0 & 0 & 0 & 0 \\
Healthy & 0 & 0 & 44 & 0 & 0 & 0 & 0 & 0 \\
Hypertension nephrosclerosis & 0 & 0 & 8 & 0 & 3 & 0 & 0 & 0 \\
Influenca vaccination & 0 & 0 & 8 & 0 & 0 & 0 & 0 & 0 \\
JIA & 0 & 0 & 85 & 0 & 0 & 0 & 0 & 0 \\
Lupus (SLE) & 0 & 0 & 2 & 0 & 1 & 0 & 0 & 0 \\
Lyme disease & 0 & 0 & 84 & 0 & 0 & 0 & 0 & 0 \\
MDS & 0 & 1 & 8 & 0 & 1 & 0 & 0 & 13 \\
MPGN-membranoproliferative?typ & 0 & 0 & 2 & 0 & 1 & 0 & 0 & 0 \\
Nephritis & 0 & 0 & 2 & 0 & 2 & 0 & 0 & 0 \\
Nephropathy & 0 & 0 & 2 & 0 & 1 & 0 & 0 & 0 \\
Plasmodium vivax challenge & 0 & 0 & 24 & 0 & 0 & 0 & 0 & 0 \\
Polycystic & 0 & 0 & 11 & 0 & 5 & 0 & 0 & 0 \\
RSV infection & 0 & 0 & 8 & 0 & 0 & 0 & 0 & 0 \\
T.ALL & 0 & 0 & 0 & 0 & 0 & 5 & 0 & 0 \\
TB.vaccination & 0 & 0 & 119 & 0 & 0 & 0 & 0 & 0 \\
Thin basement membrane dz & 0 & 0 & 2 & 0 & 1 & 0 & 0 & 0 \\
Unknown & 0 & 0 & 5 & 0 & 1 & 0 & 0 & 0 \\
Wegener's Granulomatosis & 0 & 0 & 2 & 0 & 1 & 0 & 0 & 0 \\
\bottomrule
\end{tabular}
\caption{Disease distribution across FGC-derived clusters in the AML case study.}
\label{tab:FGC_disease_distribution}
\end{table}

\begin{table}[htbp]
\centering
\scriptsize
\begin{tabular}{lcccccccc}
\toprule
\textbf{GSE Study} & \textbf{1} & \textbf{2} & \textbf{3} & \textbf{4} & \textbf{5} & \textbf{6} & \textbf{7} & \textbf{8} \\
\midrule
Bouquet (GSE63085) & 0 & 0 & 97 & 0 & 0 & 0 & 0 & 0 \\
Chen (GSE32874) & 0 & 0 & 20 & 0 & 0 & 0 & 0 & 0 \\
Divinge (GSE58335) & 0 & 0 & 6 & 0 & 0 & 0 & 0 & 0 \\
Dorr (GSE86884) & 0 & 0 & 65 & 0 & 31 & 0 & 0 & 0 \\
Doss (GSE63703) & 0 & 0 & 0 & 0 & 0 & 0 & 0 & 0 \\
GSE63816 & 0 & 0 & 0 & 0 & 0 & 0 & 0 & 0 \\
GSE72790 & 0 & 0 & 1 & 13 & 0 & 0 & 0 & 13 \\
GSE81259 & 0 & 0 & 0 & 0 & 0 & 0 & 0 & 0 \\
GSE85712 & 0 & 1 & 7 & 0 & 1 & 0 & 0 & 0 \\
Garzon (GSE63646) & 71 & 0 & 0 & 0 & 0 & 0 & 0 & 0 \\
Henn (GSE45735) & 0 & 0 & 0 & 0 & 0 & 0 & 0 & 0 \\
Hoek (GSE64655) & 0 & 0 & 8 & 0 & 0 & 0 & 0 & 0 \\
Hoft (GSE87186) & 0 & 0 & 119 & 0 & 0 & 0 & 0 & 0 \\
Lavallee (GSE49642) & 43 & 0 & 0 & 0 & 0 & 0 & 0 & 0 \\
Lavallee (GSE52656) & 27 & 0 & 0 & 0 & 0 & 0 & 0 & 0 \\
Lavallee (GSE62190) & 82 & 0 & 0 & 0 & 0 & 0 & 0 & 0 \\
Lavallee (GSE66917) & 22 & 0 & 0 & 0 & 0 & 0 & 0 & 0 \\
Lavallee (GSE67039) & 0 & 0 & 0 & 0 & 0 & 0 & 0 & 0 \\
Meldi (GSE61162) & 0 & 6 & 0 & 0 & 0 & 0 & 8 & 0 \\
Rojas.Pena (GSE67184) & 0 & 0 & 24 & 0 & 0 & 0 & 0 & 0 \\
Simon (GSE49601) & 1 & 0 & 0 & 1 & 0 & 10 & 0 & 0 \\
Spinella (GSE78785) & 0 & 0 & 0 & 0 & 0 & 5 & 0 & 0 \\
Wong (GSE79970) & 0 & 0 & 101 & 0 & 0 & 0 & 0 & 0 \\
\bottomrule
\end{tabular}
\caption{Distribution of GSE studies across FGC-derived clusters in the AML case study.}
\label{tab:FGC_GSE_distribution}
\end{table}

\begin{table}[ht]
\centering
\small
\begin{tabular}{|p{1.7cm}|p{6cm}|p{3.4cm}|p{2.5cm}|}
\hline
\textbf{Gene} & \textbf{Functional Category} & \textbf{Effect Interpretation} & \textbf{Linked to AML} \\
\hline
BTN3A1 & Immune Response and Cytokine Signaling & cell-of-origin effect &  \\
BTN3A2 & Immune Response and Cytokine Signaling & cell-of-origin effect &  \\
BTN3A3 & Immune Response and Cytokine Signaling & cell-of-origin effect &  \\
CD48   & Immune Response and Cytokine Signaling & cell-of-origin effect & \citep{wang2020acute, elias2014immune} \\
IL16   & Immune Response and Cytokine Signaling & cell-of-origin effect &  \\
GVINP1 & Immune Response and Cytokine Signaling & cell-of-origin effect &  \\
\hline
MORF4L2 & Chromatin and Epigenetic Regulators & disease-specific effect &  \\
H2AFY   & Chromatin and Epigenetic Regulators & disease-specific effect & \citep{hsu2022functional, hsu2019functional} \\
CHD1L   & Chromatin and Epigenetic Regulators & disease-specific effect &  \\
SETMAR  & Chromatin and Epigenetic Regulators & disease-specific effect & \citep{sirma2004set, handschuh2018gene, cristobal2012overexpression} \\
\hline
ZNF629 & Transcriptional Regulators. & disease-specific effect &  \\
BZW2   & Transcriptional Regulators.& disease-specific effect & \citep{nachmias2022ipo11} \\
HMGXB3 & Transcriptional Regulators. & disease-specific effect &  \\
SCML2  & Transcriptional Regulators. & disease-specific effect & \citep{grubach2008gene} \\
\hline
IPCEF1 & Catalytic Activity & cell-of-origin effect &  \\
ICK     & Catalytic Activity & cell-of-origin effect &  \\
PLCL2   & Catalytic Activity & cell-of-origin effect &  \\
CEPT1   & Catalytic Activity & cell-of-origin effect &  \\
UCK2    & Catalytic Activity & disease-specific effect & \citep{tian2025uridine, wu2019uridine} \\
NUDT15  & Catalytic Activity & disease-specific effect & \citep{wang2024nudt15} \\
OCRL    & Catalytic Activity & disease-specific effect &  \\
ACOT11  & Catalytic Activity & disease-specific effect & \citep{luo2016microarray} \\
CCNB1IP1 & Catalytic Activity & disease-specific effect & \citep{ersvaer2007cyclin} \\
BMP1    & Catalytic Activity & disease-specific effect & \citep{zylbersztejn2018bmp} \\
\hline
AKTIP   & Other & cell-of-origin effect &  \\
RETREG3 & Other & cell-of-origin effect &  \\
VDAC2   & Other & disease-specific effect &  \\
TWSG1   & Other & disease-specific effect &  \\
KCTD3   & Other & disease-specific effect &  \\
ANKRD28 & Other & disease-specific effect & \citep{de2011gene} \\
\hline
\end{tabular}
\caption{Gene-level summary of functional categories, effect source and literature linkage to AML.}
\label{tab:gene_level_summary}
\end{table}

\clearpage
\section{Runtime and Memory Profiling of Proximity-Based Clustering}\label{secA3}

The computation and storage of the full distance matrix scales quadratically with sample size, as the matrix contains $n^2$ entries. When represented as a \texttt{numpy} array of 32-bit floating point values (\texttt{float32}, 4 bytes per entry), the memory consumption grows rapidly and can become a computational bottleneck. For example, storing a full \texttt{float32} distance matrix for 1,000 samples requires approximately 4 MB of RAM, 10,000 samples require approximately 400 MB of RAM and scaling to 100,000 samples already requires approximately 40 GB of RAM, quickly resulting in memory overflows. 

\begin{figure}[ht]
    \centering
    \includegraphics[width=\textwidth]{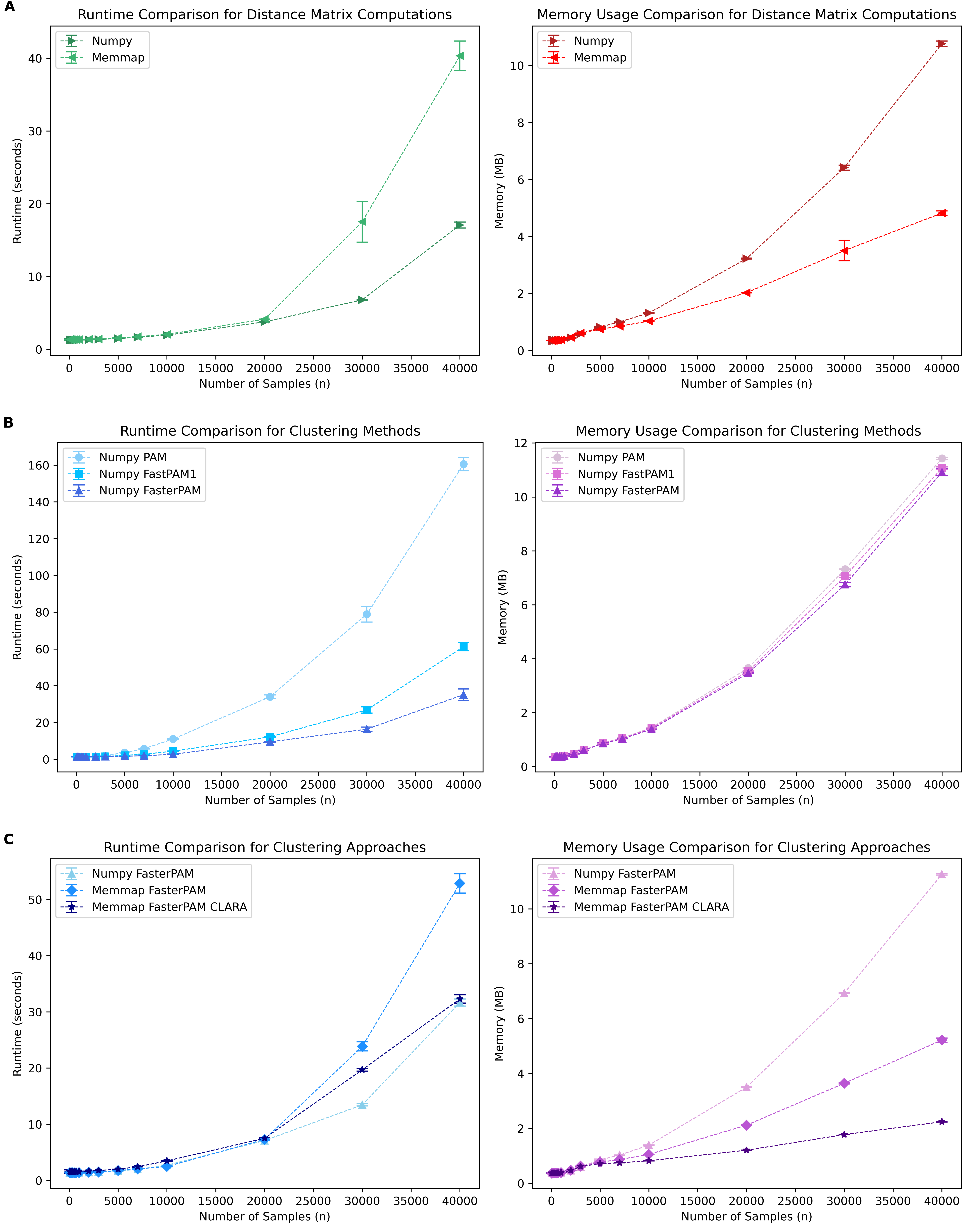}
    \caption{\textbf{Runtime and memory profiling of distance matrix computation and clustering methods.}
    (A) Runtime (left) and memory usage (right) for distance matrix computation using NumPy arrays and memory-mapped arrays (Memmap). Memmap exhibits a lower memory footprint but slightly higher runtime for large sample sizes. 
    (B) Runtime (left) and memory usage (right) of three k-medoids implementations (PAM, FastPAM1, and FasterPAM) based on NumPy distance matrices. FasterPAM is substantially faster than PAM and FastPAM1, especially for large $n$.
    (C) Combined comparison of runtime (left) and memory usage (right) for three clustering approaches: NumPy FasterPAM, Memmap FasterPAM, and Memmap FasterPAM CLARA. The CLARA-based approach achieves the lowest memory usage while maintaining efficient runtime for large-scale datasets ($n > 10{,}000$).
    Error bars denote standard deviation across ten runs. All benchmarks were performed on a MacBook Air (2024, Apple M3 chip, 24GB RAM).
    }
    \label{fig:profiling_scalability}
\end{figure}

To address these limitations, we implemented a memory-efficient backend using NumPy’s memory-mapped arrays (\texttt{memmap}), which allows on-disk storage and manipulation of large matrices without loading the entire matrix into memory. As shown in Figure~\ref{fig:profiling_scalability}A, this approach significantly improves memory efficiency while maintaining competitive runtime performance. \texttt{memmap}-based distance matrix computations demonstrate reduced memory usage across all dataset sizes, with only a moderate increase in runtime compared to standard \texttt{numpy} arrays.

For $k$-medoids clustering, we evaluated three algorithmic options provided by the \textit{kmedoids} Python package: (i) the classical \textit{PAM} algorithm (Partitioning Around Medoids) \cite{kaufman2009finding}, (ii) \textit{FastPAM1}, which reduces the cost of swap evaluations to $O(k)$ per iteration \cite{schubert2019faster}, and (iii) \textit{FasterPAM}, which builds on \textit{FastPAM1} by greedily accepting improving swaps and further reducing runtime \cite{schubert2021fast}. As shown in Figure~\ref{fig:profiling_scalability}B, \textit{FasterPAM} achieves the lowest runtime among all three variants, while maintaining similar memory usage. At $n=40{,}000$, \textit{FasterPAM} reduces clustering runtime by $\sim 20\%$ and $\sim 60\%$ compared to the \textit{FastPAM1} and the classical \textit{PAM} algorithm, respectively, without increasing memory usage. 

To further enhance scalability, we integrated \texttt{memmap} and \textit{FasterPAM} into the CLARA (Clustering Large Applications) implementation. The results, shown Figure~\ref{fig:profiling_scalability}C, demonstrate that the \textit{Memmap FasterPAM CLARA} approach provides the best trade-off between runtime and memory efficiency. At $n = 40{,}000$, CLARA reduces memory usage by more than $50\%$ compared to the standard \textit{Numpy FasterPAM} variant, with only a modest increase in runtime. This combination allows efficient clustering and feature attribution on datasets with tens of thousands of samples.

The profiling pipeline, including all benchmarking code and reproducibility scripts, is available at: \url{https://github.com/HelmholtzAI-Consultants-Munich/fg-clustering/tree/main/publication/profiling_runtime_memory}.

\end{appendices}

\clearpage
\bibliography{sn-bibliography}

\end{document}